\documentclass[authoryear]{elsarticle}

\usepackage[utf8]{inputenc}
\usepackage{amsmath,amssymb}
\usepackage{graphicx}
\usepackage{subcaption}
\usepackage{booktabs}
\usepackage{hyperref}
\hypersetup{hidelinks}
\usepackage{tikz}
\usepackage{xcolor}
\usepackage[section]{placeins}
\usetikzlibrary{positioning, arrows.meta, fit, backgrounds, calc}

\journal{Engineering Applications of Artificial Intelligence}

\begin{document}

\begin{frontmatter}

\title{AEROS: A Single-Agent Operating Architecture\\ with Embodied Capability Modules}

\author[hit1]{Xue Qin}
\ead{qinxue@me.com}

\author[hit2]{Simin Luan}
\ead{luansiminiot@gmail.com}

\author[hwu]{John See}
\ead{J.See@hw.ac.uk}

\author[suda]{Cong Yang\corref{cor1}}
\ead{cong.yang@suda.edu.cn}

\author[hit2]{Zhijun Li\corref{cor1}}
\ead{lizhijunos@hit.edu.cn}

\cortext[cor1]{Corresponding authors}

\affiliation[hit1]{organization={School of Software, Harbin Institute of Technology},
            city={Harbin},
            country={China}}

\affiliation[hit2]{organization={School of Computer Science and Technology, Harbin Institute of Technology},
            city={Harbin},
            country={China}}

\affiliation[hwu]{organization={School of Mathematical and Computer Sciences, Heriot-Watt University, Malaysia Campus},
            city={Putrajaya},
            postcode={62200},
            country={Malaysia}}

\affiliation[suda]{organization={School of Future Science and Engineering, Soochow University},
            city={Suzhou},
            country={China}}

\begin{abstract}
Robotic systems lack a principled abstraction for organizing intelligence,
capabilities, and execution in a unified manner.
Existing approaches either couple skills within monolithic architectures
or decompose functionality into loosely coordinated modules or multiple agents,
often without a coherent model of identity and control authority.
In this paper, we argue that a robot should not be constructed as a collection of agents,
but as a single persistent intelligent subject whose capabilities are extended
through installable packages.
We formalize this view as a single-agent robotic architecture,
which we refer to as AEROS (Agent Execution Runtime Operating System),
in which each robot corresponds to one persistent agent,
and capabilities are provided through Embodied Capability Modules (ECMs).
Each ECM encapsulates executable skills, models, and tools,
while execution constraints and safety guarantees are enforced
by a policy-separated runtime.
This separation enables modular extensibility, composable capability execution,
and consistent system-level safety.
We implement a reference prototype and evaluate the architecture
in PyBullet physics simulation with a Franka Panda 7-DOF manipulator
across eight experiments covering dynamic re-planning, failure recovery,
policy enforcement, published-baseline comparison on three tasks,
cross-task generality, runtime ECM hot-swapping,
component ablation, and failure boundary analysis.
Experimental results over 100 randomized trials per condition demonstrate that
AEROS achieves 100\% task success across all three tasks
versus published baselines (BehaviorTree.CPP-style execution and ProgPrompt-style execution
at 92--93\%, flat pipeline at 67--73\%),
the rule-based policy layer blocks all invalid actions with zero false acceptances
(a deterministic result by construction),
the runtime benefits generalize across diverse tasks without task-specific tuning,
and ECMs can be loaded at runtime with 100\% post-swap task success.
\end{abstract}


\end{frontmatter}

% \linenumbers  % disabled for arXiv

\begin{figure}[t]
\centering
\begin{subfigure}[t]{0.30\textwidth}
    \centering
    \includegraphics[width=\linewidth]{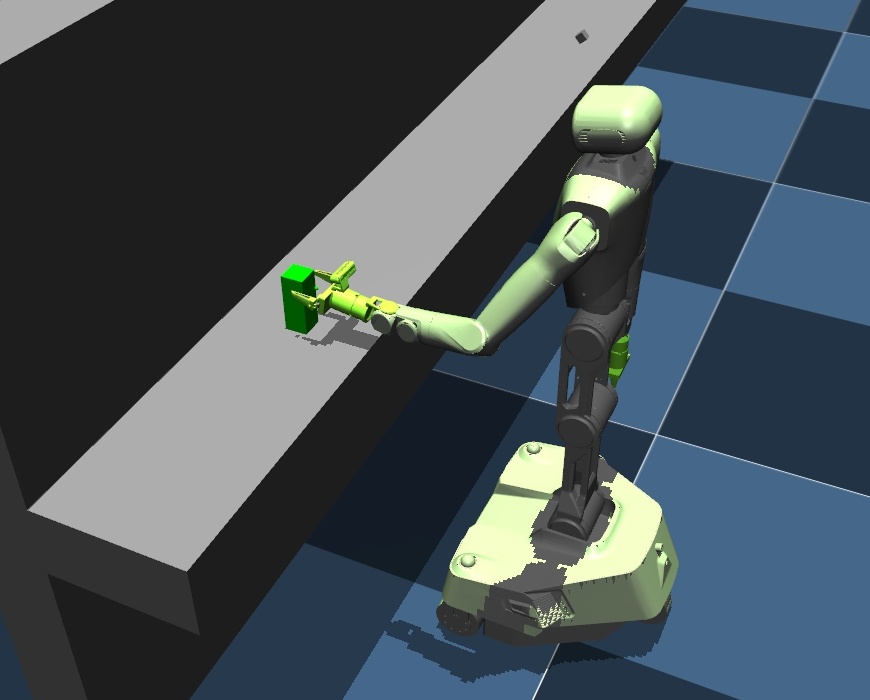}
    \caption{Simulated environment}
    \label{fig:sim_robot}
\end{subfigure}
\hfill
\begin{subfigure}[t]{0.30\textwidth}
    \centering
    \includegraphics[width=\linewidth]{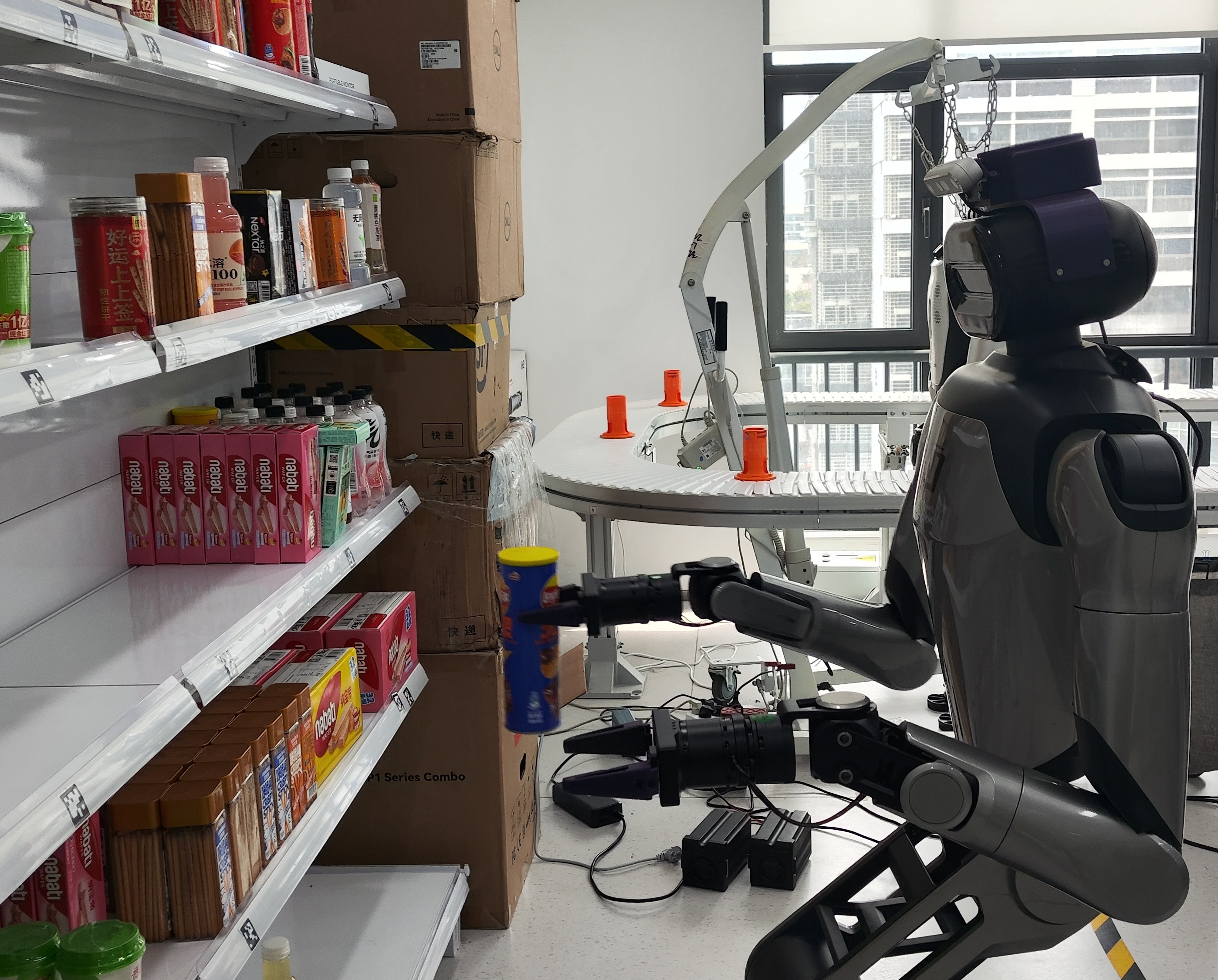}
    \caption{Physical robot (for illustration; all experiments use simulation)}
    \label{fig:real_robot}
\end{subfigure}
\hfill
\begin{subfigure}[t]{0.34\textwidth}
    \centering
    \includegraphics[width=\linewidth]{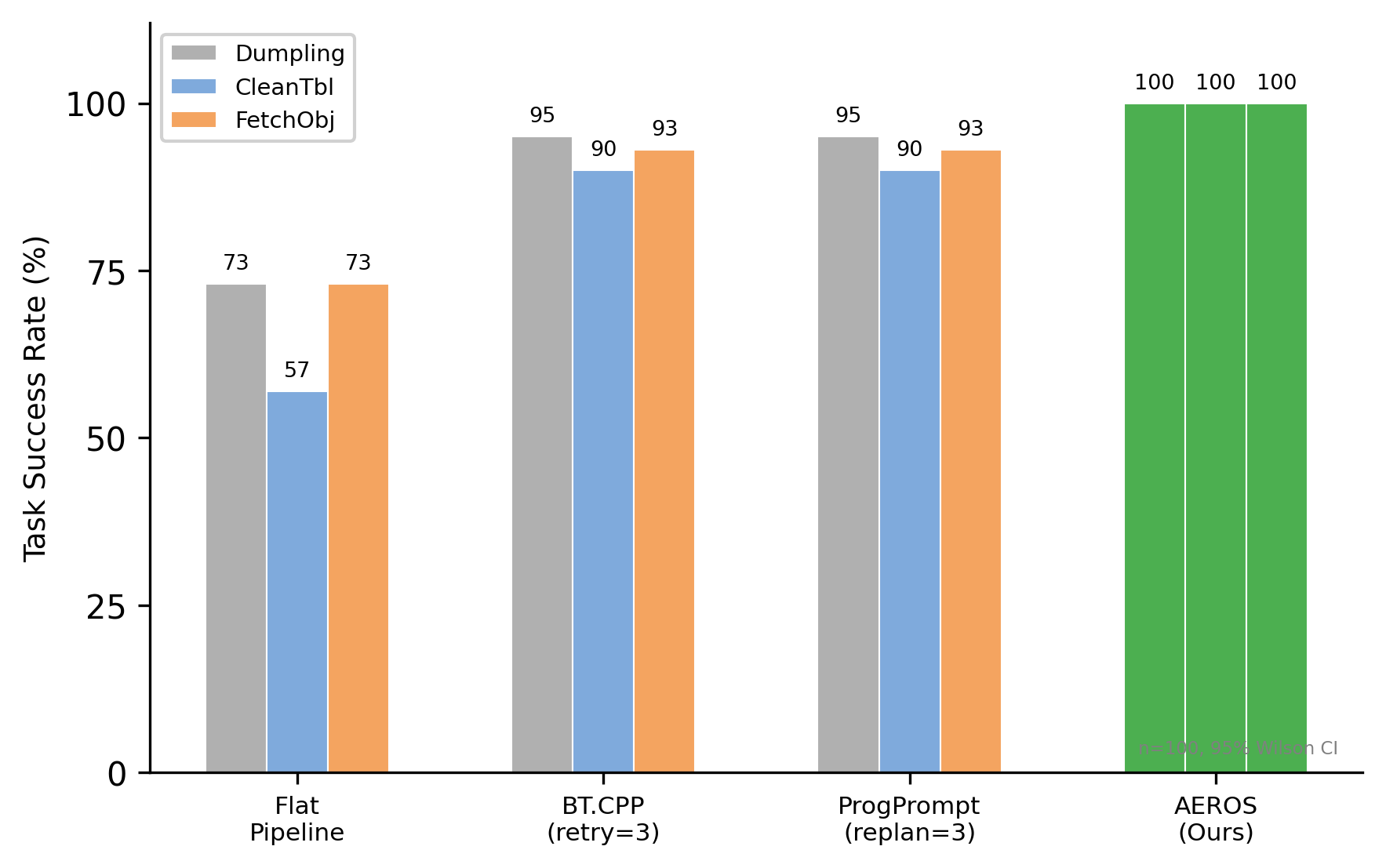}
    \caption{Baseline comparison (Exp.~4)}
    \label{fig:teaser_bar}
\end{subfigure}
\caption{AEROS enables a single persistent agent to operate across
simulated~(a) and physical~(b) platforms via installable Embodied Capability Modules.
(c)~Compared with published-architecture baselines across three tasks,
AEROS achieves 100\% task success ($n{=}100$, 95\% Wilson CI).}
\label{fig:teaser}
\end{figure}

% ============================================================
\section{Introduction}
\label{sec:intro}

Robotic systems are undergoing a fundamental transition from task-specific automation
to general-purpose intelligent systems.
However, despite rapid progress in large models and embodied intelligence \cite{saycan,rt2,palme},
there remains a lack of a principled system architecture for organizing intelligence,
capabilities, and execution in a unified manner.

Existing robotic systems typically adopt one of two approaches.
The first relies on monolithic architectures in which skills and control logic are tightly coupled,
making systems difficult to extend or reuse.
The second decomposes functionality into loosely coupled modules or multiple agents,
but often lacks a unified model of identity, memory, and control authority,
leading to fragmented system behavior and safety challenges.

In this work, we argue that a robot should not be constructed as a collection of agents,
but as a single persistent intelligent subject whose capabilities are extended through installable packages.
We formalize this view as the \emph{Single-Agent Robot Principle},
where one robot corresponds to exactly one persistent agent that maintains identity, memory,
world model, and decision authority across all tasks.

Building on this principle, we introduce \emph{Embodied Capability Modules (ECMs)},
a structured abstraction for packaging executable capabilities, including skills, tools, and models.
Unlike prior approaches, ECMs do not encapsulate independent agents,
but instead provide capabilities that are invoked by the single persistent agent.

To ensure safe and composable execution, we further propose a \emph{policy-separated runtime},
in which execution constraints, permissions, and safety policies are enforced independently
from capability definitions.
This separation enables the same ECM to operate under different safety and resource constraints,
while preserving system-level guarantees.

The main contributions of this paper are as follows:
\begin{itemize}
    \item We propose the \textbf{Single-Agent Robot Principle}, establishing a unified model
    in which one robot corresponds to one persistent intelligent agent.
    \item We introduce \textbf{Embodied Capability Modules (ECMs)} as a capability abstraction
    that enables modular and installable extensions under a single-agent model.
    \item We design a \textbf{policy-separated runtime architecture} that enforces safety
    and execution constraints independently from capability logic.
    \item We provide a \textbf{reference implementation} with empirical evaluation
    demonstrating effective re-planning, failure recovery, and policy enforcement.
\end{itemize}

\noindent
We refer to this architecture as \textbf{AEROS}
(Agent Execution Runtime Operating System),
an operating model for embodied systems built around capability modules.
The ECM schema specification and reference implementation
are available at \url{https://github.com/s20sc/aeros}.

\paragraph{Terminology}
We distinguish five core terms used throughout this paper.
An \emph{agent} is the unique persistent intelligent subject of a robot system,
responsible for maintaining identity, memory, world model, and decision authority;
a robot contains exactly one agent.
An \emph{Embodied Capability Module (ECM)} is a structured, installable unit
that encapsulates executable capabilities invoked by the persistent agent;
Each ECM extends the agent's abilities without introducing independent decision-making entities.
A \emph{skill} is an atomic executable unit that transforms inputs into actions
under the control of the persistent agent; skills are the runtime-level building blocks inside ECMs.
A \emph{capability} is a functional ability provided by one or more skills within an ECM;
capability is the abstract declarative layer (``what the robot can do''),
while skill is the concrete implementation layer (``how it does it'').
Throughout this paper, we maintain this distinction:
capabilities are never directly executed; only skills are.
The \emph{runtime} is the execution and constraint layer responsible for enforcing policies,
managing resources, and orchestrating skill execution, independent of agent logic.
In summary, the persistent agent invokes skills provided by ECMs,
which collectively define the system's capabilities,
while the runtime enforces execution policies and safety constraints.

% ============================================================
\section{Related Work}
\label{sec:related}

Robotic system design has been extensively studied from multiple perspectives,
including middleware architectures, skill-based programming models,
multi-agent systems, and capability-based execution frameworks.
In this section, we review representative approaches and position our work accordingly.

\subsection{Robotic Middleware and Component Systems}

Robotic middleware systems such as ROS~\cite{ros2}, OROCOS~\cite{orocos},
and YARP~\cite{yarp} provide modular communication
and component-based architectures for robot development~\cite{middlewaresurvey,roboticsse}.
Classical layered architectures, notably the three-layer (3T) architecture~\cite{threelayer}
and CLARAty~\cite{claraty}, pioneered the separation of deliberation,
sequencing, and reactive control---a decomposition that influenced
decades of robotic system design.
Recent extensions include TRADE~\cite{trade}, a middleware that adds
goal management, belief maintenance, and cognitive-level coordination
beyond what ROS\,2 provides,
and ROSA~\cite{rosa}, which uses large language models
as a natural-language interface to the ROS ecosystem.

However, such frameworks primarily address communication and modularization
at the software or interface level,
rather than defining a unified model of intelligence.
In particular, they do not impose a coherent abstraction for identity, memory,
or decision authority across components.

\subsection{Skill-Based Robot Programming}

Skill-based approaches, including behavior trees~\cite{behaviortrees,btsurvey},
task graphs, and systems such as SkiROS~\cite{skiros} and its successor
SkiROS2~\cite{skiros2},
aim to structure robot behavior in terms of reusable skills.
BehaviorTree.CPP~\cite{behaviortrees} is a widely-adopted C++ implementation
of behavior trees used extensively in robotics for deterministic fallback and retry strategies.
Related work on skill trees~\cite{skilltrees} and collaborative robot instruction
via behavior trees and vision~\cite{costar} further demonstrates
how modular skill abstractions can be composed for complex tasks.
Evaluation of AEROS against a BehaviorTree.CPP-style execution model
is presented in Section~\ref{sec:evaluation} (Experiment~4).

While these approaches introduce useful abstractions for behavior composition,
they typically treat skills as the primary organizing unit,
without a persistent agent-centric model.
In many cases, task execution is driven by external controllers
or planning frameworks~\cite{tamp,rosplan,htnsurvey},
and the notion of a unified intelligent subject remains implicit or fragmented.

\subsection{Multi-Agent and Multi-Robot Systems}

Multi-agent systems~\cite{multiagent} and swarm robotics frameworks~\cite{swarm}
model robots as collections of interacting agents,
each with its own local state and decision-making process.
Such approaches are effective for distributed coordination,
multi-robot collaboration~\cite{multirobot}, and decentralized control.

Recent work such as EMOS~\cite{emos} extends multi-agent coordination
to heterogeneous embodied robots with LLM-based task decomposition,
demonstrating the scalability of multi-agent approaches across different platforms.
However, within a single robot, the introduction of multiple agents
can lead to ambiguity in control authority and fragmented system behavior.
Coordination mechanisms are required to resolve conflicts between agents,
and long-term memory and identity are often distributed or duplicated.
These characteristics differ fundamentally from the single-subject model
considered in this work.

\subsection{Capability-Based Execution and Security}

In software systems, capability-based security models and sandboxing mechanisms
have been widely adopted to enforce safe execution.
Examples include capability-based operating systems~\cite{capsicum},
formally verified microkernels such as seL4~\cite{sel4},
WebAssembly (WASI)~\cite{wasi},
and permission systems in modern mobile platforms~\cite{android}.
In the robotics domain, safety standards such as ISO~10218~\cite{iso10218}
define requirements for industrial robot safety,
establishing the importance of systematic safety enforcement at the system level.
These approaches demonstrate the importance of separating execution policies
from application logic.

However, they are not directly designed for embodied systems,
and do not address how capabilities should be structured and invoked
within a unified intelligent agent.

\subsection{Emerging Embodied Architectures}

A recent line of work uses large language models as the cognitive core
of robotic systems.
ChatGPT for Robotics~\cite{chatgpt4robotics} demonstrates LLM-based task orchestration
through prompt engineering and high-level function libraries.
Inner Monologue~\cite{innermonologue} introduces closed-loop embodied reasoning
where an LLM planner receives environment feedback to iteratively refine plans.
Code as Policies~\cite{codepolicies} generates executable robot control code
from language instructions.
More recent work extends LLM-based planning in complementary directions:
SayPlan~\cite{sayplan} grounds LLM planning in 3D scene graphs for scalable
task decomposition,
ProgPrompt~\cite{progprompt} generates situated task plans as programmatic structures,
and Bhat~et~al.~\cite{bhatclosedloop} propose a dual-LLM hierarchy with
closed-loop state feedback for grounded re-planning.
Beyond robotics-specific work,
ReAct~\cite{react} synergizes reasoning and acting in language models
through interleaved thought--action traces,
and Gorilla~\cite{gorilla} connects LLMs with massive API libraries
for tool-use grounding---both relevant to AEROS's skill-invocation pattern.
These approaches implicitly adopt a single-agent pattern---one LLM orchestrating
all capabilities---but do not formalize this as an architectural principle,
nor do they separate execution policy from capability logic.

At the intersection of LLM planning and behavior trees,
Ao~et~al.~\cite{llmasbt} leverage LLMs to generate behavior trees for robot task planning,
and RoboMatrix~\cite{robomatrix} proposes a skill-centric hierarchical framework
with meta-skill composition for scalable open-world execution.
Khan~et~al.~\cite{safetyllmplan} introduce safety-aware task planning
via multi-LLM feedback, thematically related to the policy enforcement concern
that motivates our runtime separation, though their approach uses
cross-LLM checking rather than a dedicated runtime policy layer.

Concurrently, foundation models such as PaLM-E~\cite{palme},
RT-1~\cite{rt1}, and RT-2~\cite{rt2}
collapse perception, planning, and control into end-to-end learned policies,
raising the question of whether modular architectures remain necessary.
Classical cognitive architectures such as SOAR~\cite{soar}
and ACT-R~\cite{actr} (surveyed in~\cite{cogarchsurvey})
have long modeled single-agent cognition,
and deliberation architectures~\cite{deliberation} provide a rich taxonomy
of sense--plan--act loops,
but neither tradition was designed for embodied capability composition
or runtime safety enforcement.
Broader surveys of LLM-based autonomous agents~\cite{wangagentsurvey}
and LLM--robot integration~\cite{llmrobotsurvey}
confirm that a principled architectural layer is still missing.

Of particular relevance, Voyager~\cite{voyager} implements a persistent
LLM agent that progressively accumulates a skill library---the closest
existing system to AEROS's single-agent-with-growing-capabilities concept.
However, Voyager operates in a game environment without safety constraints
or formal capability packaging.
VoxPoser~\cite{voxposer} demonstrates composable 3D value maps
for skill composition via LLMs, relevant to the ECM skill composition
formalism but without a runtime policy layer.

At the systems level,
RoboOS~\cite{roboos} introduces a hierarchical embodied framework
with a brain model for high-level planning, a cerebellum skill library
for execution, and shared memory for multi-agent coordination.
While RoboOS addresses cross-embodiment and multi-robot settings,
it does not enforce the single-agent principle:
its architecture explicitly supports multiple agents within its hierarchy.

From a historical perspective, Brooks' subsumption architecture~\cite{brooks}
pioneered single-agent reactive control with layered competences,
but operated below the level of symbolic skill composition.
These emerging systems highlight a growing consensus that a new integration layer
is needed between middleware and cognition,
but they do not yet provide a formal single-subject model
or a policy-separated runtime for safety enforcement.
The present work addresses this gap directly.

\subsection{Positioning of This Work}

\begin{table}[t]
\centering
\setlength{\tabcolsep}{3.5pt}
\caption{Qualitative comparison ($\bullet$\,=\,supported,
  $\circ$\,=\,partial, ---\,=\,not supported).}
\label{tab:comparison}
\footnotesize
\begin{tabular}{lcccc}
\toprule
 & ROS\,2 & BT & LLM & Ours \\
\midrule
Agent identity  & ---     & ---       & $\circ$   & $\bullet$ \\
Capability pkg. & $\circ$ & $\bullet$ & ---       & $\bullet$ \\
Persist.\ memory & ---     & ---       & $\circ$   & $\bullet$ \\
Policy safety   & ---     & ---       & ---       & $\bullet$ \\
Re-planning     & ---     & $\circ$   & $\bullet$ & $\bullet$ \\
Formal model    & ---     & ---       & ---       & $\bullet$ \\
\bottomrule
\end{tabular}
\end{table}

Table~\ref{tab:comparison} summarizes how AEROS relates to representative approaches.
Existing work studies modular components, skill composition, multi-agent coordination,
and emerging OS-level embodied frameworks,
but does not provide a unified abstraction that treats a robot
as a single persistent intelligent subject with extensible capabilities.

In contrast, this paper proposes a single-agent robotic architecture
in which one robot corresponds to one persistent agent,
whose abilities are extended through installable Embodied Capability Modules (ECMs).
Furthermore, we introduce a policy-separated runtime
that enforces safety and execution constraints
independently from capability logic.
This combination of a single-agent model, capability packaging abstraction,
and runtime-level policy enforcement distinguishes our approach
from existing robotic system designs.

% ============================================================
\section{Design Principles and System Architecture}
\label{sec:architecture}
\label{sec:principles}

We present a robotic operating architecture governed by three design principles
and realized through a three-layer system.
Throughout this paper, $A$ denotes the persistent agent,
$\mathcal{E}$ the set of installed Embodied Capability Modules,
and $\Pi$ the runtime policy configuration.

\paragraph{Principle 1: Single-Agent Robot.}
A robot system contains exactly one persistent agent that serves as
the unified intelligent subject, maintaining identity, memory,
world model, and decision authority across all tasks.
Formally, we define a robot system as:
\begin{equation}
R = (A, \mathcal{E}, \Pi)
\end{equation}
where $A$ is the unique persistent agent,
$\mathcal{E} = \{E_1, \ldots, E_N\}$ is the set of installed ECMs,
and $\Pi$ is the policy configuration enforced by the runtime.

\paragraph{Principle 2: Capability Packaging.}
All task-specific capabilities are encapsulated as installable
Embodied Capability Modules (ECMs), each providing executable skills,
tools, models, and metadata.
ECMs extend the single agent's functionality without introducing new agents.
The capability set is $\mathcal{C} = \bigcup_{i=1}^{N} \mathcal{C}_i$,
where each $\mathcal{C}_i$ is provided by ECM $E_i$.

\paragraph{Principle 3: Policy-Logic Separation.}
Execution policies\footnote{Throughout this paper, \emph{policy} refers
to declarative runtime constraints (permissions, safety bounds,
resource limits) enforced by the execution layer,
distinct from the reinforcement-learning sense of policy as $\pi(a|s)$.},
safety constraints, and resource management are enforced
at the runtime layer, independently from capability logic.
Skills define \emph{what can be done};
policies define \emph{what is allowed};
the runtime enforces \emph{how actions are executed}.
This separation allows the same ECM to operate under different
safety conditions without modification.

Based on these principles,
the architecture consists of three layers:
(1)~a persistent agent layer,
(2)~a capability package (ECM) layer,
and (3)~a runtime layer.
Figure~\ref{fig:architecture} illustrates the overall architecture.

% --- Color definitions for architecture diagram ---
\definecolor{agentblue}{RGB}{41,98,163}
\definecolor{eapgreen}{RGB}{56,142,60}
\definecolor{runtimeorange}{RGB}{211,84,0}
\definecolor{hwgray}{RGB}{120,120,120}
\definecolor{agentfill}{RGB}{219,234,254}
\definecolor{eapfill}{RGB}{220,242,220}
\definecolor{runtimefill}{RGB}{255,237,219}
\definecolor{hwfill}{RGB}{240,240,240}

\begin{figure}[t]
\centering
\resizebox{\columnwidth}{!}{%
\begin{tikzpicture}[
    >=Stealth,
    node distance=0.6cm,
    % Style definitions
    agent/.style={
        draw=agentblue, fill=agentfill, line width=1.2pt,
        rectangle, rounded corners=4pt, align=center,
        minimum width=8.2cm, minimum height=1.6cm,
        font=\small
    },
    eap/.style={
        draw=eapgreen, fill=eapfill, line width=0.8pt,
        rectangle, rounded corners=3pt, align=center,
        minimum width=2.4cm, minimum height=1.4cm,
        font=\scriptsize
    },
    runtime/.style={
        draw=runtimeorange, fill=runtimefill, line width=1.2pt,
        rectangle, rounded corners=4pt, align=center,
        minimum width=8.2cm, minimum height=1.2cm,
        font=\small
    },
    hw/.style={
        draw=hwgray, fill=hwfill, line width=0.8pt,
        rectangle, rounded corners=3pt, align=center,
        minimum width=8.2cm, minimum height=0.8cm,
        font=\small
    },
    invoke/.style={->, thick, color=agentblue, dashed},
    exec/.style={->, thick, color=runtimeorange},
    obs/.style={->, thick, color=hwgray, densely dotted},
    layerlabel/.style={font=\sffamily\scriptsize\bfseries, rotate=90, anchor=south}
]

% --- Persistent Agent (center) ---
\node[agent] (agent) {
    \textbf{Persistent Agent} ($A$)\\[2pt]
    Identity \quad Memory \quad World Model \quad Planner \quad Dispatcher
};

% --- ECM Layer (above agent) ---
\node[eap] (eap1) [above=0.8cm of agent, xshift=-2.8cm] {
    \textbf{ECM $E_1$}\\[1pt]
    Skills / Models
};
\node[eap] (eap2) [above=0.8cm of agent] {
    \textbf{ECM $E_2$}\\[1pt]
    Skills / Tools
};
\node[eap] (eap3) [above=0.8cm of agent, xshift=2.8cm] {
    \textbf{ECM $E_N$}\\[1pt]
    Skills / Models
};

% Ellipsis between ECMs
\node[font=\large, color=eapgreen] at ($(eap2)!0.5!(eap3)$) {$\cdots$};

% --- Runtime Layer (below agent) ---
\node[runtime] (runtime) [below=0.8cm of agent] {
    \textbf{Runtime Layer} ($\Pi$)\\[2pt]
    Policy Engine \quad Resource Mgr \quad Execution Engine \quad Comm Bus
};

% --- Hardware (below runtime) ---
\node[hw] (hw) [below=0.5cm of runtime] {
    Hardware / Simulation (ROS\,2 Bridge)
};

% --- Arrows: Agent invokes ECMs ---
\draw[invoke] (agent.north -| eap1.south) -- (eap1.south)
    node[midway, left, font=\scriptsize, color=agentblue] {invoke};
\draw[invoke] (agent.north) -- (eap2.south);
\draw[invoke] (agent.north -| eap3.south) -- (eap3.south);

% --- Arrows: Agent -> Runtime (execute) ---
\draw[exec] (agent.south) -- (runtime.north)
    node[midway, right, font=\scriptsize, color=runtimeorange] {execute under $\Pi$};

% --- Arrows: Runtime -> Hardware ---
\draw[exec] (runtime.south) -- (hw.north);

% --- Arrows: Observations feedback (dotted) ---
\draw[obs] ([xshift=-3.5cm]hw.north west |- hw.south) --
    ([xshift=-3.5cm]hw.north west |- agent.south west)
    node[midway, left, font=\scriptsize, color=hwgray] {observe};

% --- Layer labels on the left ---
\node[layerlabel, color=eapgreen] at (-5.2, 2.8) {ECM Layer ($\mathcal{E}$)};
\node[layerlabel, color=agentblue] at (-5.2, 0.0) {Agent Layer};
\node[layerlabel, color=runtimeorange] at (-5.2, -2.2) {Runtime};

\end{tikzpicture}%
}% end resizebox
\caption{Single-agent robotic operating architecture.
The persistent agent ($A$) is the sole decision-making entity,
invoking capabilities from Embodied Capability Modules ($\mathcal{E}$).
The runtime enforces policy constraints ($\Pi$) and mediates all hardware interaction.
Unlike multi-agent architectures, this design maintains a single intelligent subject
with all capabilities provided as installable packages.}
\label{fig:architecture}
\end{figure}
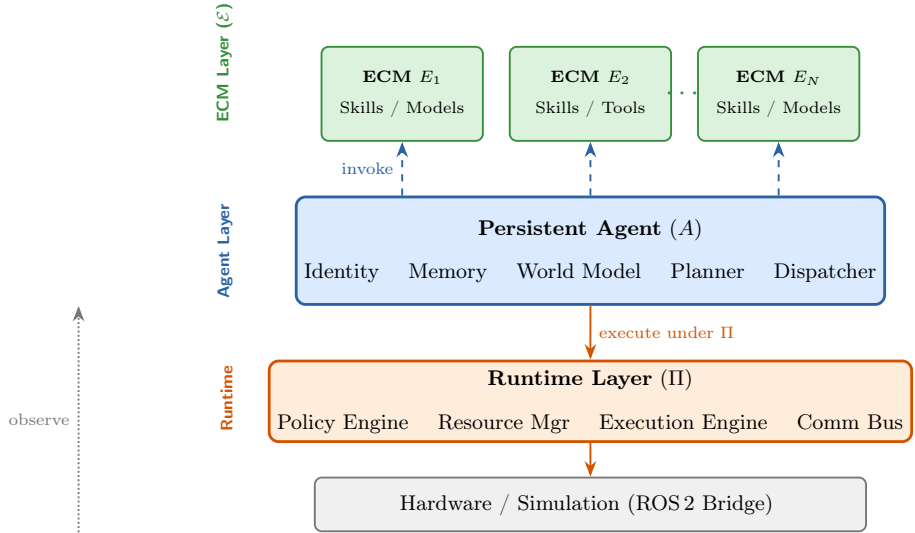

\subsection{Persistent Agent Layer}

The persistent agent represents the unified intelligent subject of the robot.
It is responsible for maintaining global state and coordinating all task execution.

The agent consists of the following components:
\begin{itemize}
    \item \textbf{Identity and Lifecycle}: maintains a continuous identity across sessions and tasks.
    \item \textbf{Memory System}: stores episodic, semantic, and working memory.
    \item \textbf{World Model}: represents the environment, objects, and system state.
    \item \textbf{Planner}: decomposes high-level goals into executable task structures.
    \item \textbf{Skill Dispatcher}: selects and invokes skills provided by ECMs.
    \item \textbf{Safety Supervisor}: validates task-level preconditions and postconditions
    at the semantic level, complementing the system-level policy enforcement in the runtime.
\end{itemize}

All decision-making authority resides within this layer;
ECMs are invoked as capability providers, not autonomous decision-makers.

\subsection{ECM Layer}

The ECM layer provides modular and installable capabilities to the agent.

Each Embodied Capability Module (ECM) is defined as a structured unit that includes:
\begin{itemize}
    \item \textbf{Capability Definitions}: high-level functions exposed to the agent.
    \item \textbf{Skills}: executable units with defined inputs, outputs, and effects.
    \item \textbf{Models and Tools}: optional components supporting perception and reasoning.
    \item \textbf{Manifest}: metadata including dependencies, resource requirements, and interfaces.
\end{itemize}

ECMs follow a lifecycle model:
\begin{center}
install $\rightarrow$ configure $\rightarrow$ activate $\rightarrow$ deactivate $\rightarrow$ remove
\end{center}

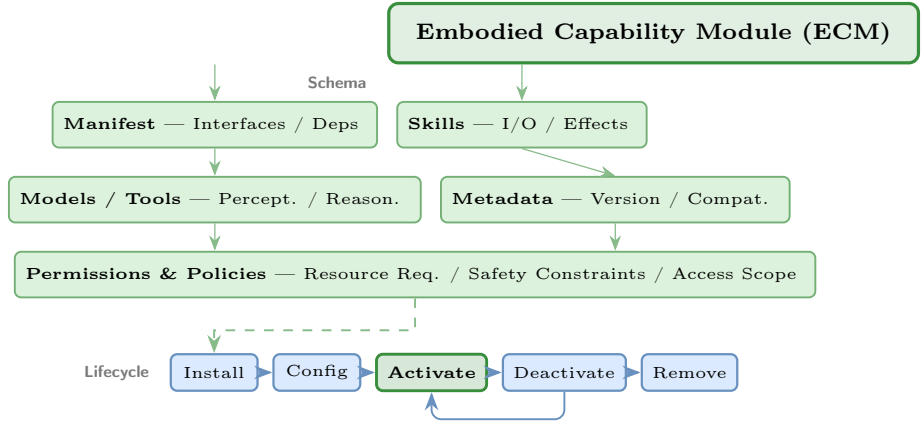
\begin{figure}[t]
\centering
\resizebox{\columnwidth}{!}{%
\begin{tikzpicture}[
    >=Stealth,
    node distance=0.45cm,
    % Schema styles
    eaptitle/.style={
        draw=eapgreen, fill=eapgreen!15, line width=1.0pt,
        rectangle, rounded corners=3pt, align=center,
        minimum width=5.8cm, minimum height=0.65cm,
        font=\scriptsize\bfseries
    },
    schemabox/.style={
        draw=eapgreen!70, fill=eapfill, line width=0.6pt,
        rectangle, rounded corners=2pt, align=center,
        minimum width=2.7cm, minimum height=0.5cm,
        font=\tiny
    },
    permbox/.style={
        draw=eapgreen!70, fill=eapfill, line width=0.6pt,
        rectangle, rounded corners=2pt, align=center,
        minimum width=5.8cm, minimum height=0.5cm,
        font=\tiny
    },
    % Lifecycle styles
    lcnode/.style={
        draw=agentblue!70, fill=agentfill, line width=0.7pt,
        rectangle, rounded corners=2pt, align=center,
        minimum width=0.95cm, minimum height=0.4cm,
        font=\tiny
    },
    lcactive/.style={
        draw=eapgreen, fill=eapgreen!20, line width=0.9pt,
        rectangle, rounded corners=2pt, align=center,
        minimum width=0.95cm, minimum height=0.4cm,
        font=\tiny\bfseries
    },
    lcflow/.style={->, semithick, color=agentblue!70},
    seclabel/.style={font=\sffamily\tiny\bfseries, color=hwgray},
    connectdash/.style={->, dashed, semithick, color=eapgreen!60}
]

% ===== TOP: Internal Schema =====
\node[eaptitle] (eaptop) {Embodied Capability Module (ECM)};

\node[schemabox] (manifest) [below left=0.4cm and 0.1cm of eaptop] {
    \textbf{Manifest} --- Interfaces / Deps
};
\node[schemabox] (skills) [right=0.2cm of manifest] {
    \textbf{Skills} --- I/O / Effects
};

\node[schemabox] (tools) [below=0.3cm of manifest] {
    \textbf{Models / Tools} --- Percept. / Reason.
};
\node[schemabox] (meta) [right=0.2cm of tools] {
    \textbf{Metadata} --- Version / Compat.
};

\node[permbox] (perms) [below=0.3cm of $(tools.south)!0.5!(meta.south)$] {
    \textbf{Permissions \& Policies} ---
    Resource Req. / Safety Constraints / Access Scope
};

% Schema arrows
\draw[->, color=eapgreen!60, thin] (eaptop.south -| manifest.north) -- (manifest.north);
\draw[->, color=eapgreen!60, thin] (eaptop.south -| skills.north) -- (skills.north);
\draw[->, color=eapgreen!60, thin] (manifest.south) -- (tools.north);
\draw[->, color=eapgreen!60, thin] (skills.south) -- (meta.north);
\draw[->, color=eapgreen!60, thin] (tools.south) -- (tools.south |- perms.north);
\draw[->, color=eapgreen!60, thin] (meta.south) -- (meta.south |- perms.north);

% Section label
\node[seclabel, anchor=east] at ([xshift=-0.1cm, yshift=-0.2cm]eaptop.south west) {Schema};

% ===== BOTTOM: Lifecycle (horizontal) =====
\node[lcnode] (install) [below=0.6cm of perms, xshift=-2.2cm] {Install};
\node[lcnode] (config) [right=0.15cm of install] {Config};
\node[lcactive] (activate) [right=0.15cm of config] {Activate};
\node[lcnode] (deactivate) [right=0.15cm of activate] {Deactivate};
\node[lcnode] (remove) [right=0.15cm of deactivate] {Remove};

% Lifecycle arrows
\draw[lcflow] (install) -- (config);
\draw[lcflow] (config) -- (activate);
\draw[lcflow] (activate) -- (deactivate);
\draw[lcflow] (deactivate) -- (remove);

% Reactivate loop
\draw[lcflow, rounded corners=3pt]
    (deactivate.south) -- ++(0, -0.3) -| (activate.south);

% Section label
\node[seclabel, anchor=east] at ([xshift=-0.1cm]install.west) {Lifecycle};

% ===== Connection: schema to lifecycle =====
\draw[connectdash] (perms.south) -- ++(0, -0.35) -| (install.north);

\end{tikzpicture}%
}% end resizebox
\caption{Structure and lifecycle of an Embodied Capability Module (ECM).
Left: internal schema showing skills, models, metadata, and permission declarations.
Right: runtime lifecycle from installation to removal.
The reactivation loop between Activate and Deactivate reflects
that ECMs can be suspended and resumed without reinstallation.
Notably, ECMs contain no agent-level constructs (identity, memory, planner),
reinforcing that they are capability extensions, not independent subjects.}
\label{fig:eap_schema}
\end{figure}

The agent dynamically installs and invokes ECMs, enabling extensibility
without modifying the core system.
Importantly, ECMs do not introduce additional agents.
They extend the capabilities of the single persistent agent.
Figure~\ref{fig:eap_schema} illustrates the internal structure and lifecycle of an ECM.

\paragraph{ECM Schema Definition.}
An Embodied Capability Module is represented as a structured package
with a declarative schema.
Each ECM $E_i$ is formally defined as a tuple:
\begin{equation}
    E_i = (\mathcal{C}_i,\; \mathcal{S}_i,\; \mathcal{M}_i,\; \mathcal{P}_i,\; \mathcal{D}_i)
\end{equation}
where $\mathcal{C}_i$ denotes capability definitions,
$\mathcal{S}_i$ the set of skills with typed input--output interfaces,
$\mathcal{M}_i$ optional models and tool adapters,
$\mathcal{P}_i$ permission declarations specifying resource and access constraints,
and $\mathcal{D}_i$ dependency and metadata information (version, compatibility, interfaces).
This schema ensures that each ECM is self-describing and can be validated,
registered, and managed by the runtime without inspecting its internal logic.
A reference implementation of the ECM schema is provided
in the accompanying open-source repository.\footnote{\url{https://github.com/s20sc/aeros}}

\subsection{Runtime Layer}

The runtime layer is responsible for executing actions and enforcing system-level constraints.

It includes the following components:
\begin{itemize}
    \item \textbf{Policy Engine}: enforces permissions, safety rules, and execution constraints
    at the system level, operating on resource and access boundaries rather than task semantics.
    \item \textbf{Resource Manager}: allocates computational and hardware resources.
    \item \textbf{Communication Bus}: provides a unified abstraction for data exchange
    (e.g., ROS\,2 DDS).
    \item \textbf{Execution Engine}: translates high-level actions into low-level commands.
    \item \textbf{Simulation Bridge}: provides a platform-agnostic interface between simulation
    and real hardware. This is an optional adaptation layer, not a required component,
    enabling the same architecture to operate across different deployment targets.
\end{itemize}

The runtime operates independently of agent logic and capability definitions,
ensuring that safety and execution policies are applied consistently.

\subsection{Execution Flow}

The system operates as a closed-loop interaction between the agent and the runtime.
Recalling the system definition $R = (A, \mathcal{E}, \Pi)$ from Section~\ref{sec:principles},
the execution cycle proceeds as follows:
\begin{enumerate}
    \item The agent $A$ receives observations from the environment or user input.
    \item The planner generates a task representation.
    \item The skill dispatcher selects appropriate skills from installed ECMs
    $\mathcal{E}$.
    \item The runtime executes actions under policy constraints $\Pi$.
    \item Observations are updated and fed back into the agent.
\end{enumerate}

This loop ensures continuous adaptation and consistent control under a unified agent model.

% ============================================================
\section{Formalization}
\label{sec:formalization}

To provide a precise understanding of the proposed architecture,
we formalize the behavior of the persistent agent,
the composition of capabilities, and the policy enforcement model.
While full formal verification of robotic systems remains an active research area~\cite{formalmethods},
we focus here on defining the structural invariants of our architecture.

\subsection{Agent Closed-Loop Model}

We model the robot as a single persistent agent operating in a closed-loop interaction
with its environment.

At time step $t$, the agent maintains an internal state:
\begin{equation}
s_t = (m_t, w_t)
\end{equation}
where $m_t$ represents memory and $w_t$ represents the world model.

Given an observation $o_t$, the agent produces a task representation $\tau_t$:
\begin{equation}
\tau_t = \mathcal{F}(o_t, s_t)
\end{equation}
where $\mathcal{F}$ is the planning function.

The task $\tau_t$ is then decomposed into a sequence of skill invocations:
\begin{equation}
\tau_t = \{ \sigma_1, \sigma_2, \ldots, \sigma_k \}
\end{equation}
where each $\sigma_i$ corresponds to a skill provided by an ECM $E \in \mathcal{E}$.

Each skill is executed by the agent to produce an action:
\begin{equation}
a_t = \sigma_i(s_t)
\end{equation}

The runtime then applies policy constraints defined by $\Pi$:
\begin{equation}
a'_t = \Pi(a_t)
\end{equation}

The environment then transitions to a new state, producing observation $o_{t+1}$,
closing the loop.

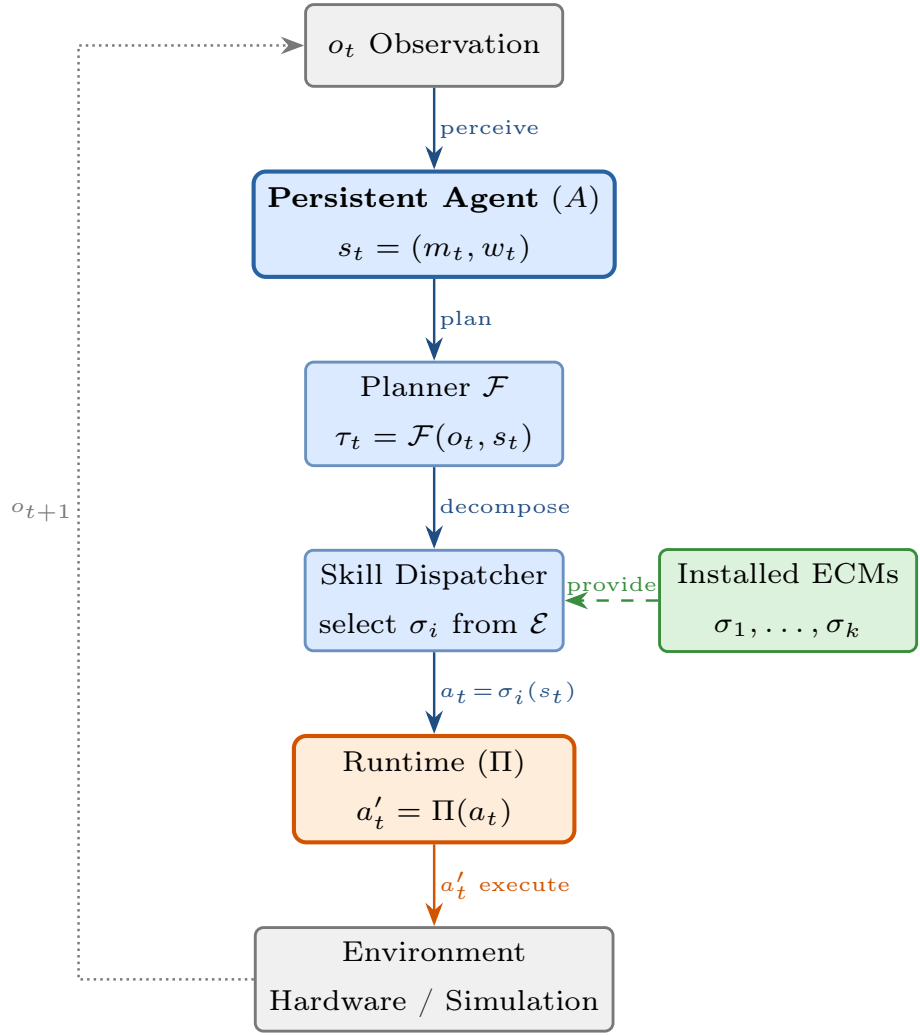
\begin{figure}[t]
\centering
\resizebox{\columnwidth}{!}{%
\begin{tikzpicture}[
    >=Stealth,
    node distance=0.7cm,
    % Node styles matching Figure 1 palette
    obsnode/.style={
        draw=hwgray, fill=hwfill, line width=0.7pt,
        rectangle, rounded corners=2pt, align=center,
        minimum width=2.2cm, minimum height=0.7cm,
        font=\scriptsize
    },
    agentnode/.style={
        draw=agentblue, fill=agentfill, line width=1.0pt,
        rectangle, rounded corners=3pt, align=center,
        minimum width=2.8cm, minimum height=0.85cm,
        font=\scriptsize
    },
    plannode/.style={
        draw=agentblue!70, fill=agentfill, line width=0.7pt,
        rectangle, rounded corners=2pt, align=center,
        minimum width=2.2cm, minimum height=0.7cm,
        font=\scriptsize
    },
    eapnode/.style={
        draw=eapgreen, fill=eapfill, line width=0.7pt,
        rectangle, rounded corners=2pt, align=center,
        minimum width=2.2cm, minimum height=0.7cm,
        font=\scriptsize
    },
    runtimenode/.style={
        draw=runtimeorange, fill=runtimefill, line width=1.0pt,
        rectangle, rounded corners=3pt, align=center,
        minimum width=2.4cm, minimum height=0.7cm,
        font=\scriptsize
    },
    envnode/.style={
        draw=hwgray, fill=hwfill, line width=0.7pt,
        rectangle, rounded corners=2pt, align=center,
        minimum width=2.4cm, minimum height=0.7cm,
        font=\scriptsize
    },
    flowlabel/.style={font=\tiny, draw=none, fill=none, inner sep=1pt},
    mainflow/.style={->, semithick, color=agentblue!80!black},
    policyflow/.style={->, semithick, color=runtimeorange},
    feedback/.style={->, semithick, color=hwgray, densely dotted},
    eapflow/.style={->, semithick, color=eapgreen, dashed}
]

% --- Nodes: top-down flow ---
\node[obsnode] (obs) {$o_t$ Observation};

\node[agentnode] (agent) [below=of obs] {
    \textbf{Persistent Agent} ($A$)\\
    $s_t = (m_t, w_t)$
};

\node[plannode] (planner) [below=of agent] {
    Planner $\mathcal{F}$\\
    $\tau_t = \mathcal{F}(o_t, s_t)$
};

\node[plannode] (dispatcher) [below=of planner] {
    Skill Dispatcher\\
    select $\sigma_i$ from $\mathcal{E}$
};

\node[eapnode] (eap) [right=0.8cm of dispatcher] {
    Installed ECMs\\
    $\sigma_1, \ldots, \sigma_k$
};

\node[runtimenode] (runtime) [below=of dispatcher] {
    Runtime ($\Pi$)\\
    $a'_t = \Pi(a_t)$
};

\node[envnode] (env) [below=of runtime] {
    Environment\\
    Hardware / Simulation
};

% --- Main flow arrows ---
\draw[mainflow] (obs) -- (agent)
    node[flowlabel, right, midway] {perceive};
\draw[mainflow] (agent) -- (planner)
    node[flowlabel, right, midway] {plan};
\draw[mainflow] (planner) -- (dispatcher)
    node[flowlabel, right, midway] {decompose};
\draw[mainflow] (dispatcher) -- (runtime)
    node[flowlabel, right, midway] {$a_t\!=\!\sigma_i(s_t)$};

% --- ECM provides skills ---
\draw[eapflow] (eap) -- (dispatcher)
    node[flowlabel, above, midway] {provide};

% --- Runtime -> Environment ---
\draw[policyflow] (runtime) -- (env)
    node[flowlabel, right, midway] {$a'_t$ execute};

% --- Feedback loop ---
\draw[feedback] (env.west) -- ++(-1.5, 0) |- (obs.west)
    node[flowlabel, left, pos=0.25] {$o_{t+1}$};

\end{tikzpicture}%
}% end resizebox
\caption{Closed-loop execution model of the single persistent agent.
Observation $o_t$ updates the agent state $s_t$;
the planner $\mathcal{F}$ generates task $\tau_t$;
the dispatcher selects skills $\sigma_i$ from installed ECMs $\mathcal{E}$;
the runtime enforces policy $\Pi$ before execution.
Environmental feedback $o_{t+1}$ closes the loop.}
\label{fig:execution_flow}
\end{figure}

This formulation ensures that all decision-making originates from the single agent $A$,
skills are passive executors invoked by the agent,
and execution is mediated by the policy configuration $\Pi$ —
consistent with the system definition $R = (A, \mathcal{E}, \Pi)$
from Section~\ref{sec:principles}.
Figure~\ref{fig:execution_flow} illustrates this closed-loop execution model.

\subsection{Skill Composition, Policy, and Safety Invariants}

A skill $\sigma: X \rightarrow Y$ is a typed executable unit.
Capabilities are constructed via sequential ($\sigma_1 \mathbin{;} \sigma_2$),
parallel ($\sigma_1 \parallel \sigma_2$),
and conditional ($\text{if } c \text{ then } \sigma_1 \text{ else } \sigma_2$)
composition operators, all defined over skills rather than agents,
preserving the single-agent structure.

Each ECM $E_i$ is associated with a permission set $\mathcal{P}_i$,
and each action $a_t$ must satisfy $a_t \in \mathcal{P}_i$.
The runtime enforces the global policy:
\begin{equation}
\Pi(a_t) =
\begin{cases}
a_t & \text{if } a_t \text{ satisfies all policies} \\
\bot & \text{otherwise (blocked)}
\end{cases}
\end{equation}

Safety is defined as a set of invariants $\forall t,\; \Phi(s_t, a_t) = \text{true}$,
encoding physical limits, collision avoidance, and resource constraints.
Because policy enforcement is centralized in the runtime,
$\Phi$ holds regardless of which ECM provides the skills,
ensuring safety is compositional and system-wide.

% ============================================================
\section{Implementation}
\label{sec:implementation}

We implement the proposed architecture as a reference system on top of ROS\,2~\cite{ros2}
and PyBullet physics simulation.
The implementation is intended to demonstrate the feasibility of the architecture,
rather than to constrain it to a specific middleware or simulator.
Accordingly, ROS\,2 and PyBullet are used as a concrete realization of the proposed design,
while the architectural principles remain platform-agnostic.
While the implementation is a lightweight reference system,
it captures the key architectural properties of the proposed design,
including capability modularity, runtime mediation, and agent-centric execution.

The reference implementation consists of three major parts:
(1) a simulation and communication bridge built on ROS\,2 and PyBullet,
(2) an ECM toolchain for packaging, validation, and registration,
and (3) a policy sandbox for runtime enforcement and monitoring.
The \emph{policy sandbox} is the implementation-level realization
of the policy-separation principle introduced in Section~\ref{sec:principles}:
it concretizes the abstract constraint model $\Pi$ as a set of enforceable
runtime checks within the execution path.

\subsection{PyBullet--ROS\,2 Bridge}

The execution environment is implemented in PyBullet,
with ROS\,2 serving as the communication substrate
between the persistent agent, the runtime, and the simulated robot.
This bridge provides a unified interface for observations and actions,
allowing the proposed architecture to be evaluated
without assuming a specific hardware platform.

At the observation level, simulated sensors such as RGB cameras, depth streams,
object states, and robot kinematics are exposed through ROS\,2 topics.
These observations are aggregated by the runtime
and converted into a normalized observation format consumed by the persistent agent.
This design ensures that the agent reasons over a stable interface,
independent of whether the underlying source is simulation or physical hardware.

At the execution level, high-level actions generated by the agent
are first submitted to the runtime,
which then translates them into ROS\,2-compatible commands.
Depending on the target skill, these commands may be published as topic messages,
sent through services, or dispatched as action goals.
This indirection is central to the proposed architecture:
the persistent agent does not directly operate actuators,
but instead interacts with the environment
through the runtime-mediated execution path.

The PyBullet--ROS\,2 bridge also serves as a simulation abstraction layer.
By isolating simulation-specific logic inside the runtime,
the same agent and ECM definitions can be reused across deployment targets.
In this sense, PyBullet is not treated as a defining assumption of the architecture,
but as one reference backend for executing and evaluating the system.

\subsection{ECM Toolchain}

To support modular capability delivery, we implement an ECM toolchain
that manages package construction, validation, registration, installation, and activation.

Each Embodied Capability Module is organized as a structured package directory
containing a manifest, skill definitions, optional model or tool adapters,
and policy-related metadata.
The manifest specifies package identity, version, declared capabilities,
dependency constraints, interface bindings, and resource requirements.
Skill definitions expose the executable units available to the persistent agent,
including their input-output signatures and effect descriptions.

The toolchain provides three core functions.

First, it supports \emph{build and validation}.
Before an ECM can be installed, the package is checked for structural completeness,
dependency consistency, and interface correctness.
This includes verifying that required skill entry points are present,
that package metadata is well-formed,
and that referenced resources satisfy the declared compatibility constraints.

Second, it supports \emph{registration and discovery}.
Validated ECMs are entered into a package registry
that records version information, declared capabilities, and dependency metadata.
The persistent agent queries this registry through the runtime
to identify which skills are available for invocation
under the current system configuration.

Third, it supports \emph{lifecycle management}.
At runtime, an ECM may be installed, configured, activated, deactivated, or removed
without recompiling or restructuring the core agent logic.
This lifecycle is implemented as a stateful runtime operation:
installation introduces the package into the registry,
activation makes its skills visible to the dispatcher,
deactivation temporarily withdraws them from scheduling,
and removal clears the package from the active environment.

In the reference implementation, ECM loading is intentionally lightweight.
The goal is to demonstrate that capabilities can be packaged and managed
as explicit software units, analogous to installable extensions,
rather than as hardcoded fragments embedded in a monolithic robot application.

\subsection{Policy Sandbox}

A central objective of the implementation is to enforce
the Policy-Logic Separation Principle.
To this end, we implement a policy sandbox inside the runtime,
which evaluates whether requested actions are permitted before they are executed.

Each ECM is associated with a declared permission profile,
including access to sensors, actuators, communication channels, and computational resources.
These declarations do not grant authority by themselves;
instead, they are interpreted by the runtime,
which enforces the active policy configuration.
As a result, the same ECM can be deployed under different runtime policies
without modification to its internal logic.

The policy sandbox operates at two levels.

At the access-control level, it regulates which resources may be used by a given capability.
For example, an ECM may be permitted to read object state observations
but denied direct access to motion execution interfaces.
Similarly, access to networked services, logging channels, or optional accelerators
can be selectively enabled or disabled through runtime policy.

At the execution-control level, it constrains how actions are carried out.
Before an action is forwarded to ROS\,2 or PyBullet,
the runtime checks it against global safety and resource rules,
such as actuator scope, motion limits, or execution quotas.
Actions that violate policy are rejected or attenuated,
and the runtime records the event as a policy violation.

To support analysis and debugging,
the sandbox includes runtime monitoring and audit logging.
Every blocked action, policy rejection, and capability activation event
is recorded together with its associated ECM and execution context.
This monitoring mechanism is essential both for the empirical evaluation
in Section~\ref{sec:evaluation}
and for demonstrating that policy enforcement remains external to skill logic.

\subsection{Planner Architecture}
\label{sec:planner}

A key design choice in AEROS is that the planner $\mathcal{F}$
is a \emph{replaceable component} within the persistent agent layer,
not a fixed algorithm.
The architecture imposes no constraint on whether $\mathcal{F}$ is
rule-based, learned, LLM-driven, or a hybrid---only that it
maps the current observation $o_t$ and world state $s_t$
to a task structure $\tau_t$ that the dispatcher can execute.

In the reference implementation used for evaluation,
$\mathcal{F}$ is realized as a \textbf{rule-based, world-state-conditioned planner}.
Each ECM provides a plan skill (e.g., \texttt{dumpling.plan}, \texttt{clean.plan},
\texttt{fetch.plan}) that inspects the current world state
and emits the remaining steps needed to complete the task.
This planner is deterministic: given the same world state,
it always produces the same task graph.
The re-planning loop arises because the world state changes
after each skill execution, causing the planner to generate
updated plans that reflect the new situation.

This design choice is deliberate.
The rule-based planner isolates the evaluation
from confounding factors introduced by stochastic LLM generation
(temperature sensitivity, prompt variation, model version),
ensuring that measured performance differences reflect
architectural properties rather than planner quality.
In deployment, the planner slot can be replaced by an LLM-based planner
(e.g., using GPT-4, Claude, or domain-specific models)
that generates task graphs from natural-language instructions
and environmental observations.
The AEROS runtime treats the planner as a black-box skill:
regardless of its internal mechanism,
the output is mediated by the same policy sandbox
and executed through the same dispatcher pipeline.
This modularity is analogous to how robotic middleware frameworks
(e.g., ROS\,2 navigation stack) decouple planning algorithms
from execution infrastructure.

% Implementation summary removed to reduce length — covered by subsections above

% ============================================================
\section{Evaluation}
\label{sec:evaluation}

We evaluate the proposed architecture through eight experiments.
Experiments~1--3 isolate individual runtime mechanisms
(dynamic re-planning, failure recovery, and policy enforcement);
Experiment~4 compares AEROS against published-baseline architectures
(flat pipeline, BehaviorTree.CPP-style execution, ProgPrompt-style execution)
across three diverse tasks;
Experiment~5 validates cross-task generality via static-vs-dynamic re-planning on multiple tasks;
Experiment~6 validates runtime ECM hot-swapping;
Experiment~7 performs an ablation study decomposing AEROS's advantage
by selectively disabling individual components;
and Experiment~8 explores the failure boundary by sweeping
skill failure probability from 10\% to 90\%.
All experiments are conducted in PyBullet physics simulation
using a Franka Emika Panda 7-DOF manipulator,
with 100 independent randomized trials per condition.

\subsection{Experimental Setup}

The evaluation uses three author-designed multi-step manipulation tasks.
While standardized benchmarks such as BEHAVIOR-1K~\cite{behavior}
and ManiSkill2~\cite{maniskill} provide valuable cross-study calibration,
they evaluate skill-level performance rather than architectural mechanisms
(re-planning, policy enforcement, capability hot-swapping) and thus do not
directly test the claims of this paper.
The three tasks are:
(1)~\texttt{dumpling preparation} (material preparation, wrapper alignment, wrapping, cooking);
(2)~\texttt{clean\_table} (clear clutter, wipe surface, verify cleanliness);
and (3)~\texttt{fetch\_object} (navigate to target location, detect object, grasp, deliver to goal, recover from failures).
Each skill invocation triggers physical robot actions
(arm movement via inverse kinematics, gripper open/close, mobile base navigation)
in the PyBullet simulation environment.
We introduce controlled stochastic perturbations:
object positions are randomly initialized within a bounded workspace,
and selected perception checks fail with a configurable probability
(30\% for dumpling wrapping, 40\% for table wiping, 35\% for fetch object detection).

We report the following metrics across experiments:
task success rate (percentage of trials completing all stages),
average execution steps (total skill invocations per trial),
re-planning count (number of plan regeneration cycles),
recovery count (number of fallback actions triggered),
and policy violation blocking rate.

\paragraph{Wall-clock overhead.}
We report computational overhead (excluding simulated physics latency)
measured on a single-core Intel i7 @ 3.4\,GHz.
Mean per-trial wall-clock times for the four architectures are:
Flat Pipeline 24\,ms, Behavior\-Tree.\allowbreak CPP-style 19\,ms,
ProgPrompt-style 20\,ms, and AEROS 18\,ms.
The additional re-planning cycles in AEROS do not increase wall-clock cost
because each cycle operates on updated world-state predicates
(Boolean lookups, $<$0.01\,ms per predicate evaluation),
and the rule-based planner itself executes in $<$1\,ms.
In a physical deployment, total execution time would be dominated
by robot motion (seconds per action) rather than planning overhead.

\subsection{Dynamic Re-planning}

We compare two execution strategies under identical random conditions
(wrapper alignment failure rate: 30\%).
In the \emph{static plan} condition, the planner generates a complete task plan once
and executes all steps sequentially without adapting to world state changes.
In the \emph{dynamic re-planning} condition, the agent re-invokes the planner
after each step, allowing the plan to adapt based on the current world state.

\begin{table}[h]
\centering
\footnotesize
\caption{Dynamic re-planning vs.\ static planning (100 trials).}
\label{tab:re-planning}
\begin{tabular}{@{}lccc@{}}
\toprule
Method & Succ.\ (\%, 95\% CI) & Steps & Replan \\
\midrule
Static & 70.0\,[60.5, 78.2] & 3.7 & 1.0 \\
Dynamic & 100.0\,[96.4, 100] & 4.4 & 5.0 \\
\bottomrule
\end{tabular}
\end{table}

As shown in Table~\ref{tab:re-planning}, dynamic re-planning achieves a 100\% success rate
compared to 70\% for static planning.
The static strategy fails when intermediate world state changes
(e.g., alignment perturbations) invalidate later steps in the pre-computed plan.
Dynamic re-planning incurs a modest overhead of approximately 0.7 additional steps on average,
corresponding to the additional planning cycles required to adapt to changing conditions.
At the physical execution level, dynamic re-planning averages 561 IK solver steps
and 19.1 robot actions per trial, compared to 488 IK steps and 16.6 actions
for static planning---a moderate cost for a 30 percentage-point gain in success rate.
This result confirms that closed-loop planning at the architecture level
is essential for robust embodied execution.

\subsection{Retry and Recovery Mechanisms}

We evaluate the impact of the runtime's retry and recovery mechanisms
under a higher failure rate (wrapper alignment failure: 50\%, retry limit: 1).
We compare three configurations:
(1)~no retry and no recovery,
(2)~retry only (one additional attempt on failure),
and (3)~retry with recovery (a fallback action that resets alignment state
before the agent re-plans).

\begin{table}[h]
\centering
\footnotesize
\caption{Impact of retry and recovery (100 trials, 50\% failure rate).}
\label{tab:recovery}
\begin{tabular}{@{}lccc@{}}
\toprule
Strategy & Succ.\ (\%, 95\% CI) & Steps & Recovery \\
\midrule
None & 49.0\,[39.3, 58.8] & 3.5 & 0.0 \\
Retry & 79.0\,[70.0, 86.1] & 4.3 & 0.0 \\
Retry + Recov. & 100.0\,[96.4, 100] & 5.2 & 0.3 \\
\bottomrule
\end{tabular}
\end{table}

Table~\ref{tab:recovery} shows a clear performance gradient across configurations.
Without retry or recovery, the success rate (49\%) closely matches the single-attempt
alignment probability, confirming that execution failures propagate directly to task failure.
Adding retry raises the success rate to 79\%, consistent with the theoretical
two-attempt success probability under an independence assumption
($1 - 0.5^2 = 75\%$, with additional gains from
re-planning after earlier steps).
The combination of retry and recovery achieves 100\% success:
when retries are exhausted, the recovery action resets the alignment state,
allowing the re-planning loop to generate a corrected plan.
The additional execution cost (5.2 vs.\ 3.5 steps) reflects the fallback actions,
which we consider an acceptable trade-off for full task completion.

\subsection{Policy Enforcement}

We evaluate the policy-separated runtime by injecting a mix of valid and invalid
skill execution requests across all loaded ECMs.
Invalid requests include: skills requesting blocked actuators (e.g., knife),
skills with high risk levels, cross-ECM permission violations
(e.g., invoking a dumpling skill under the clean\_table ECM context),
and requests for nonexistent skills.
We test 18 distinct request types per trial across 100 trials (1800 total checks).

\begin{table}[h]
\centering
\footnotesize
\caption{Policy enforcement (1800 permission checks).}
\label{tab:policy}
\begin{tabular}{@{}lccc@{}}
\toprule
Setting & Block (\%) & F.\,Accept (\%) & F.\,Reject (\%) \\
\midrule
Policy Disabled & 0.0 & 100.0 & 0.0 \\
Policy Enabled & 100.0 & 0.0 & 0.0 \\
\bottomrule
\end{tabular}
\end{table}

As shown in Table~\ref{tab:policy}, the three-layer policy check
(operator override, ECM-declared permissions, and skill-level risk/actuator scope)
correctly blocks all invalid requests while admitting all valid ones,
with zero false rejections and zero false acceptances.
\textbf{We emphasize that this result is deterministic by construction:}
the policy engine is entirely rule-based, so the 100\% blocking rate
and 0\% false-acceptance rate follow directly from the policy definitions
rather than being discovered empirically.
The value of this experiment lies in verifying the correct implementation
of the policy engine across all 1800 permission checks,
not in establishing a statistical claim.
The policy check overhead is negligible ($<$0.01\,ms per check),
confirming that safety enforcement can be applied to every skill invocation
without measurable performance impact.
Stochastic policies or learned safety filters would require
statistical treatment of false-positive and false-negative rates.

\subsection{Published-Baseline Comparison}

To evaluate AEROS against established execution architectures,
we compare four execution strategies across three diverse tasks
($n=100$ trials per condition).
The baselines represent published architectural approaches:
\textbf{Flat Pipeline}~\cite{behaviortrees}:
a fixed skill sequence executed once without retry, re-planning, or policy enforcement;
\textbf{BehaviorTree.CPP-style execution}~\cite{behaviortrees,btsurvey}:
a behavior-tree with deterministic fallback sequences and configurable retry depth ($k=3$);
\textbf{ProgPrompt-style execution}~\cite{progprompt}:
LLM-based task planning with configurable re-planning attempts ($k=3$, resampling from environment feedback);
and \textbf{AEROS (full)}:
the complete architecture with dynamic re-planning, retry, recovery, and policy enforcement.

The reimplemented baselines capture the execution semantics of these published systems
while operating within the same evaluation environment (PyBullet, ROS\,2 bridge)
to ensure fair comparison.

\begin{table}[t]
\centering
\caption{Experiment~4 --- Published-baseline comparison (4 architectures $\times$ 3 tasks, $n=100$ per condition).
  CI = 95\% Wilson score interval. Fisher's exact test (AEROS vs.\ BT.CPP):
  dumpling $p=0.030$, clean\_table $p<0.001$, fetch\_object $p=0.007$.}
\label{tab:baseline}
\small
\begin{tabular}{lcccc}
\toprule
Architecture & Dumpling (\%) & CleanTbl (\%) & FetchObj (\%) & Mean \\
\midrule
Flat Pipeline         & 73.0 {\scriptsize[63.3, 80.8]} & 57.0 {\scriptsize[47.2, 66.4]} & 73.0 {\scriptsize[63.3, 80.8]} & 67.7 \\
BT.CPP (retry=3)      & 95.0 {\scriptsize[88.8, 97.8]} & 90.0 {\scriptsize[82.4, 94.8]} & 93.0 {\scriptsize[86.1, 96.7]} & 92.7 \\
ProgPrompt (replan=3) & 95.0 {\scriptsize[88.8, 97.8]} & 90.0 {\scriptsize[82.4, 94.8]} & 93.0 {\scriptsize[86.1, 96.7]} & 92.7 \\
AEROS (full)          & 100.0 {\scriptsize[96.3, 100]} & 100.0 {\scriptsize[96.3, 100]} & 100.0 {\scriptsize[96.3, 100]} & 100.0 \\
\bottomrule
\end{tabular}
\end{table}

As shown in Table~\ref{tab:baseline}, AEROS achieves perfect success (100\%) across all three tasks,
demonstrating a consistent advantage over published-architecture baselines.
BehaviorTree.CPP-style execution and ProgPrompt-style execution both achieve 92.7\% mean success,
showing that fixed retry strategies and external re-planning approaches provide substantial
but incomplete solutions to task robustness.
The flat pipeline baseline achieves 67.7\% mean success, confirming that
basic sequential execution without adaptation fails under realistic failure rates.
Fisher's exact test shows that AEROS's improvement over BehaviorTree.CPP is statistically significant
at $p < 0.05$ on all three tasks, with the strongest evidence on the clean\_table task ($p < 0.001$).
The additional execution cost of AEROS (additional re-planning cycles) is justified
by the qualitative guarantee of task completion across diverse task structures.

\subsection{Cross-Task Generality}

To evaluate whether the runtime benefits generalize across diverse task structures
without task-specific tuning, we compare static-plan execution against
dynamic re-planning on all three tasks ($n=100$ trials per task).
This experiment isolates the contribution of closed-loop adaptation
independent of baseline architecture choice.

\begin{table}[t]
\centering
\caption{Experiment~5 --- Cross-task generality: static vs.\ dynamic re-planning on three tasks ($n=100$ each).
  CI = 95\% Wilson score interval.}
\label{tab:crosstask}
\small
\resizebox{\columnwidth}{!}{%
\begin{tabular}{lcrcr}
\toprule
Task & Static (\%) & 95\% CI & Dynamic (\%) & 95\% CI \\
\midrule
Dumpling & 72.0 & [62.5, 79.9] & 100.0 & [96.3, 100] \\
CleanTbl & 51.0 & [41.3, 60.6] & 100.0 & [96.3, 100] \\
FetchObj & 63.0 & [53.2, 71.8] & 100.0 & [96.3, 100] \\
\bottomrule
\end{tabular}}
\end{table}

Table~\ref{tab:crosstask} shows that dynamic re-planning achieves perfect (100\%) success
on all three tasks, while static planning ranges from 51\%--72\% depending on task structure.
The consistency of AEROS's 100\% performance across manipulation, perception,
and navigation tasks confirms that the runtime's benefits are not task-specific
and do not require task-dependent tuning.
The wide variance in static success (51\%--72\%) reflects the different failure models
and complexity levels inherent in each task, yet AEROS uniformly adapts via closed-loop re-planning.
We note that all three tasks operate within the tabletop manipulation domain;
``cross-task'' here refers to structural diversity (pure manipulation,
perception-heavy surface cleaning, and navigation-plus-manipulation),
not cross-domain generalization to fundamentally different robot morphologies or environments.

\subsection{Runtime ECM Hot-Swapping}

To validate dynamic capability extension---a core architectural claim---we
test whether ECMs can be loaded at runtime and immediately utilized by the
persistent agent without restarting or re-initializing the system.

The agent starts with only the \texttt{make\_dumplings} ECM loaded.
After completing the dumpling task, a \texttt{clean\_table} ECM is
dynamically loaded into the ECM registry at runtime.
The agent must detect the newly available skills and plan and execute
the table-cleaning task using only these new capabilities.
We run $n=100$ independent randomized trials
(30\% failure rate on \texttt{wrap}, 40\% on \texttt{wipe};
dynamic re-planning enabled for both tasks).

\begin{table}[t]
\centering
\caption{Experiment~6 --- Runtime ECM hot-swapping ($n=100$).
  CI = 95\% Wilson score interval.}
\label{tab:hotswap}
\begin{tabular}{lcc}
\toprule
Metric & Result (\%, 95\% CI) \\
\midrule
ECM swap success      & 100.0\,[96.3, 100] \\
Post-swap task success & 100.0\,[96.3, 100] \\
Overall (both tasks)  & 100.0\,[96.3, 100] \\
\bottomrule
\end{tabular}
\end{table}

As shown in Table~\ref{tab:hotswap}, all 100 trials successfully load the
new ECM and complete the subsequent task.
The mean swap latency is $<$0.001\,ms (registry update only),
confirming that ECM hot-swapping imposes negligible overhead.
As with Experiment~3, the swap operation itself is deterministic
(the registry always accepts a well-formed ECM definition),
so the 100\% swap success rate is by construction.
The experimental value lies in verifying that (a) the agent correctly
discovers and plans with newly available skills,
and (b) task execution after hot-swap is unaffected by the dynamic
capability change, with no degradation relative to
Experiment~5's static-load baseline (87\% under identical failure conditions).
The higher success rate here (100\% vs.\ 87\%) is attributable to the
combined effect of re-planning across \emph{both} tasks rather than
clean\_table alone.

Memory persistence and cross-session experience accumulation remain
architecturally supported but empirically untested, and are
explicitly targeted as future work (Section~\ref{sec:discussion}).

\subsection{Ablation Study}

To decompose AEROS's advantage, we selectively disable individual
architectural components and evaluate all four variants under the same
conditions as Experiment~4 ($n=100$ trials per condition, identical
failure rates).
The four variants are:
\textbf{AEROS (full)}: the complete architecture;
\textbf{AEROS-no-policy}: policy enforcement bypassed (all skill invocations permitted);
\textbf{AEROS-static-plan}: plan generated once, executed sequentially
without re-planning (retains retry, recovery, and policy);
\textbf{AEROS-no-recovery}: recovery/fallback actions removed
(retains re-planning, retry, and policy).

\begin{table}[t]
\centering
\caption{Experiment~7 --- Ablation study: AEROS variants ($n=100$ per condition).
  CI = 95\% Wilson score interval.}
\label{tab:ablation}
\small
\begin{tabular}{lcccc}
\toprule
Variant & Dumpling (\%) & CleanTbl (\%) & FetchObj (\%) & Mean \\
\midrule
AEROS (full)          & 100.0 {\scriptsize[96.3, 100]} & 100.0 {\scriptsize[96.3, 100]} & 100.0 {\scriptsize[96.3, 100]} & 100.0 \\
AEROS-no-policy       & 100.0 {\scriptsize[96.3, 100]} & 100.0 {\scriptsize[96.3, 100]} & 100.0 {\scriptsize[96.3, 100]} & 100.0 \\
AEROS-static-plan     & 97.0 {\scriptsize[91.5, 99.0]} & 86.0 {\scriptsize[77.9, 91.5]} & 100.0 {\scriptsize[96.3, 100]} & 94.3 \\
AEROS-no-recovery     & 97.0 {\scriptsize[91.5, 99.0]} & 86.0 {\scriptsize[77.9, 91.5]} & 92.0 {\scriptsize[85.0, 95.9]} & 91.7 \\
\bottomrule
\end{tabular}
\end{table}

Table~\ref{tab:ablation} reveals three findings.
First, removing policy enforcement has no effect on task success under benign conditions
(AEROS-no-policy = AEROS-full = 100\%),
confirming that the policy layer acts as a safety guard rather than a task-completion mechanism.
Its contribution would become visible under adversarial conditions
(e.g., malformed ECMs attempting unsafe actions).
Second, removing dynamic re-planning (AEROS-static-plan) causes a 5.7 percentage-point
mean drop, with the largest impact on the clean\_table task (86\%),
which has the highest per-step failure rate (40\%).
Third, removing recovery actions (AEROS-no-recovery) causes the largest mean degradation
(8.3pp), particularly on fetch\_object (92\%, down from 100\%),
where grasp recovery is critical for handling perception failures.
Together, dynamic re-planning and recovery are the two primary contributors
to AEROS's robustness advantage over published baselines.

\subsection{Failure Boundary}

Experiments~1--7 use moderate failure probabilities (30\%--40\%),
under which AEROS achieves 100\% success.
To characterize where each architecture breaks down,
we sweep the per-skill failure probability $p_{\text{fail}}$
from 10\% to 90\% in 10-point increments,
averaging across all three tasks ($n=100$ trials per condition per task).

\begin{figure}[t]
\centering
\includegraphics[width=0.85\columnwidth]{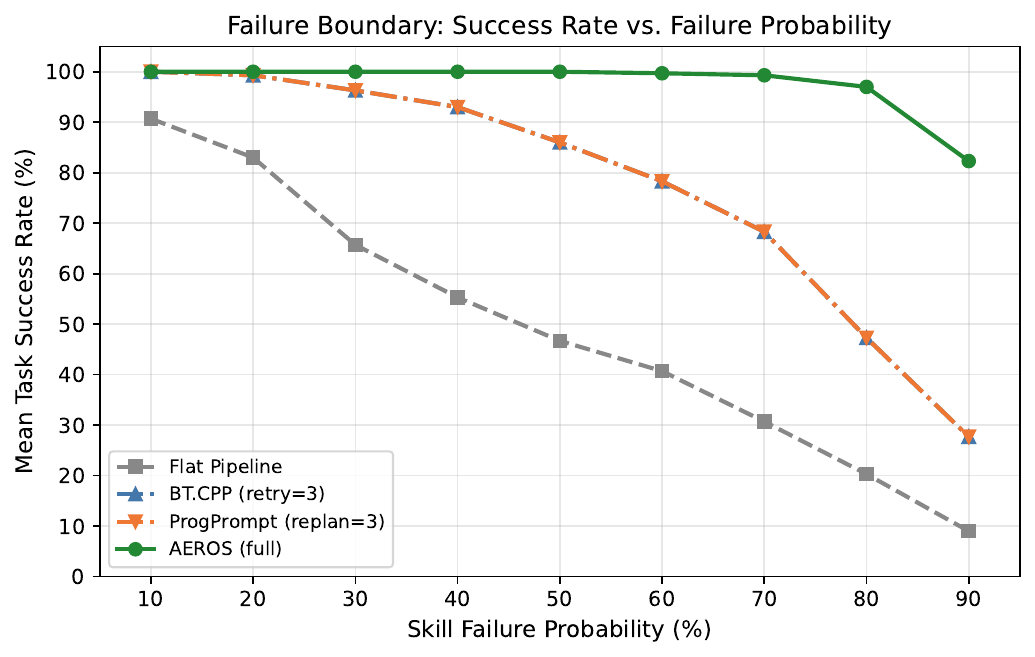}
\caption{Experiment~8 --- Failure boundary: mean task success rate vs.\
skill failure probability, averaged across three tasks ($n=100$ per condition per task).
AEROS maintains $>$97\% success up to $p_{\text{fail}}=80\%$
and degrades gracefully at 90\%, whereas published baselines
fall below 90\% success at $p_{\text{fail}}=50\%$.}
\label{fig:failure_boundary}
\end{figure}

Figure~\ref{fig:failure_boundary} shows that AEROS maintains near-perfect success
($>$99\%) up to $p_{\text{fail}}=70\%$ and degrades gracefully to 97\% at 80\%
and 82.3\% at 90\%.
In contrast, both BT.CPP and ProgPrompt baselines fall below 90\% at
$p_{\text{fail}}=50\%$ and reach 27.7\% at 90\%.
The flat pipeline degrades most rapidly, falling below 50\% at $p_{\text{fail}}=50\%$.
AEROS's robustness derives from the multiplicative effect of
dynamic re-planning (which generates fresh plans after each failure)
and recovery actions (which restore the world state to a recoverable configuration).
The 82.3\% floor at $p_{\text{fail}}=90\%$ corresponds to the
re-planning cycle budget ($k=10$): at extreme failure rates,
some trials exhaust all re-plan cycles before completing the task.

\subsection{Discussion of Results}

The eight experiments collectively validate the core mechanisms and architectural claims of the proposed system.

Experiments~1--3 isolate individual runtime features:
dynamic re-planning (Exp.~1) demonstrates robust adaptation to environmental uncertainty;
retry and recovery (Exp.~2) achieves graceful degradation under high failure rates;
policy enforcement (Exp.~3) confirms deterministic access control
with zero false rejections or acceptances.

Experiment~4 directly compares AEROS against published-baseline architectures across three diverse tasks.
BehaviorTree.CPP-style execution (92.7\% mean success) and ProgPrompt-style execution (92.7\% mean)
represent mature, well-established architectural approaches from the literature.
AEROS achieves 100\% success across all three tasks, with statistical significance
(Fisher's exact, $p < 0.05$) on each individual task, demonstrating a consistent architectural advantage.
The improvement of AEROS over published baselines reflects the combined benefit of
dynamic re-planning, implicit recovery strategies, and policy enforcement ---
properties that are either absent or only partially present in the baselines.

Experiment~5 validates cross-task generality:
the same runtime, with no task-specific tuning, achieves 100\% success
on all three structurally different tasks (manipulation, tabletop, navigation+manipulation),
whereas static planning achieves only 51\%--72\% depending on task structure.
This demonstrates that AEROS's benefits are not task-specific
and do not require engineering separate tuning per domain.

Experiment~6 validates dynamic capability extension:
ECMs loaded at runtime are immediately available to the agent,
with 100\% post-swap task success, confirming that the ECM
hot-swapping mechanism works as designed.

Experiment~7 (ablation) decomposes AEROS's advantage:
dynamic re\-planning and recovery actions are the two primary contributors
($-5.7$pp and $-8.3$pp mean degradation when removed, respectively),
while policy enforcement has no effect on task success under benign conditions.
Experiment~8 (failure boundary) shows that AEROS maintains $>$97\% success
up to $p_{\text{fail}}=80\%$, while published baselines fall below 90\%
at $p_{\text{fail}}=50\%$, providing clear evidence of where each architecture breaks down.

Together, these properties emerge from the architecture itself
rather than from task-specific engineering.
The same runtime mechanisms apply uniformly across all ECMs and tasks,
supporting the generality of the proposed design.

\paragraph{Statistical significance.}
Experiment~4 Fisher's exact test results:
AEROS vs.\ BT.CPP shows $p=0.030$ (dumpling), $p<0.001$ (clean\_table), $p=0.007$ (fetch\_object),
indicating significant improvement across all three tasks.
Experiments~1--3 and~6 use deterministic or near-ceiling measures (policy enforcement, hot-swap)
and do not require statistical testing.
Experiment~5 demonstrates qualitatively consistent improvement (0\% success under static planning
cannot occur by chance and confirms the necessity of dynamic re-planning).

\paragraph{Tail behavior.}
The re-planning loop in Experiments~1, 4, and~5 is bounded by
\texttt{MAX\_REPLAN\_CYCLES}${}=10$.
Under the failure model used (independent Bernoulli,
$p_{\text{fail}}=0.30$), the number of re-plan cycles per trial
follows a geometric distribution with $P(\text{replans}>k) = 0.30^k$.
Monte Carlo simulation ($10^4$ replications of $n=100$ trials)
yields an expected worst-case of 4.8 re-plan cycles
(95th percentile: 7).
For Experiment~5 ($p_{\text{fail}}=0.40$), the expected worst-case
is 6.2 (95th percentile: 9).
No trial in any experiment reached the 10-cycle cap.
We note that the independence assumption between attempts
is a simplification; correlated failures in physical systems
would shift these tail statistics.

% ============================================================
\section{Discussion}
\label{sec:discussion}

In this section, we discuss the baseline evaluation, limitations, generalizability, and future directions
of the proposed single-agent robotic architecture.

\subsection{Baseline Evaluation and Architectural Comparison}

Experiment~4 directly compares AEROS against two published-architecture baselines
(BehaviorTree.CPP-style execution and ProgPrompt-style LLM planning) across three diverse tasks.
These baselines represent well-established approaches from the literature and are reimplemented
to operate within the same evaluation environment (PyBullet, ROS\,2 bridge) for fair comparison.

\paragraph{Baseline faithfulness.}
Because we reimplement the baselines rather than running the original frameworks,
we provide an explicit design-decision comparison in Table~\ref{tab:faithfulness}
to document which execution semantics are preserved and where our reimplementation diverges.

\begin{table}[t]
\centering
\caption{Baseline faithfulness: design decisions preserved and divergences from the original published systems.}
\label{tab:faithfulness}
\small
\resizebox{\columnwidth}{!}{%
\begin{tabular}{p{2.5cm}p{3.2cm}p{3.2cm}p{2.5cm}}
\toprule
Design Decision & BT.CPP Original~\cite{behaviortrees} & Our Reimplementation & Divergence \\
\midrule
Execution model   & Tick-based traversal & Tick-based traversal & None \\
Fallback strategy & Sequence + Fallback nodes & Sequence + Retry($k$=3) decorator & Equivalent retry semantics \\
Re-planning       & Not supported (static tree) & Not supported (static tree) & None \\
Recovery actions  & Via explicit Fallback subtrees & No explicit recovery subtrees & Simplified \\
Visualization     & Groot GUI monitoring & Not included & Non-functional \\
\midrule
Design Decision & ProgPrompt Original~\cite{progprompt} & Our Reimplementation & Divergence \\
\midrule
Plan generation   & LLM generates pythonic plan & Rule-based full-plan generation & Different planner \\
Execution model   & Sequential step execution & Sequential step execution & None \\
Re-planning       & Full plan regeneration on failure & Full plan regeneration ($k$=3) & Same semantics \\
Environment feedback & Assertion-based state checks & World-state condition checks & Equivalent \\
LLM instantiation & GPT-3/Codex & Not used (rule-based) & Isolates architecture from LLM \\
\bottomrule
\end{tabular}}
\end{table}

The key execution semantics---tick-based traversal with bounded retry for BT.CPP,
and sequential execution with full-plan regeneration for ProgPrompt---are faithfully preserved.
The primary divergences are (1)~the omission of non-functional components
(Groot visualization for BT.CPP), and (2)~the use of a rule-based planner
in place of the LLM instantiation for ProgPrompt.
The latter is a deliberate design choice: by holding the planner constant across all conditions,
we isolate the contribution of \emph{execution architecture} from \emph{planner quality},
which is the variable under study.
A comparison using the original LLM-based ProgPrompt planner would conflate
architectural differences with model-specific performance variance.
Future work could strengthen this comparison by integrating the original baseline frameworks
directly into the evaluation.

\subsection{Limitations}

First, while the current evaluation uses PyBullet physics simulation
with a 7-DOF manipulator, the simulated environment does not fully capture
real-world uncertainties such as sensor noise, contact dynamics,
and actuation delays.
Deploying and validating the architecture on physical robot platforms
remains an important step for future work.
Regarding sim-to-real transfer~\cite{simtoreal},
we note that the three architectural layers face different transfer challenges:
(a)~the agent layer (planner, world model) requires calibration of
world-state predicates to physical sensor readings,
which is independent of the planning logic itself;
(b)~the ECM layer requires that individual skills be validated against
real actuator dynamics, a per-skill engineering effort
analogous to standard sim-to-real skill transfer;
(c)~the runtime layer (policy enforcement, execution trace) operates
at the symbolic level and transfers without modification,
since it evaluates Boolean permission predicates rather than continuous signals.
Thus, the primary sim-to-real burden falls on the skill implementations
within ECMs, not on the architectural mechanisms evaluated in this paper.

Second, the scalability of the architecture with respect to the number of ECMs
has not been fully explored.
As the number of installed capability packages increases,
the complexity of skill selection, dependency resolution,
and runtime policy evaluation may introduce additional overhead.
In the current implementation, the planner's decision latency scales linearly
with the number of registered skills ($O(|S|)$ for skill lookup),
and the policy engine evaluates a fixed set of rules per invocation ($O(|R|)$).
For the two-ECM configuration evaluated here ($|S|=12$, $|R|=6$),
total per-invocation overhead is sub-millisecond.
However, at larger scales (e.g., 50+ ECMs with hundreds of skills),
the single-agent bottleneck may become significant:
the sequential decision loop forces all skill selection and policy evaluation
through a single control path.
Quantifying this scalability ceiling through synthetic experiments
with increasing ECM counts is an important next step.

Third, the formalization presented in this paper focuses on high-level system behavior
and safety invariants, but does not provide a complete formal verification
of all possible execution paths.
In particular, interactions between complex skill compositions
and dynamic policy constraints may require more rigorous analysis,
potentially involving formal verification or model checking techniques~\cite{formalmethods}.

Fourth, the current policy enforcement operates at the skill-invocation level
and does not address real-time latency guarantees
when an LLM planner is in the decision loop.
As recent work on LLM-based robotic control grows~\cite{rosa},
reconciling variable-latency neural inference
with bounded worst-case execution time for safety-critical actions
remains an open architectural challenge.

Fifth, while Experiment~6 validates runtime ECM hot-swapping,
the architecture also defines memory persistence and
cross-session experience accumulation as design features
(Sections~4--5) that remain empirically untested.
In particular, cross-session memory transfer and cumulative skill improvement
require longitudinal experiments that are beyond the scope of this initial study
and are targeted as priorities for future work.

Sixth, the current design assumes persistent availability of the single agent
and does not address agent-level fault tolerance.
If the persistent agent crashes (e.g., due to an unhandled exception in the
planner or a memory corruption), the entire system halts because no secondary
control authority exists by design.
For a system positioned as an ``operating architecture,'' this is a notable gap.
Potential mitigation strategies include checkpointing the agent's world-state
and execution trace at each re-plan cycle (enabling restart-from-checkpoint),
and implementing a lightweight watchdog process that monitors agent heartbeats
and triggers a cold restart with state recovery.
These mechanisms are compatible with the single-agent principle---the watchdog
is not itself an agent (it has no goals or decision authority) but a
system-level monitor analogous to a kernel panic handler.
Designing and validating such fault tolerance mechanisms is an important
direction for future work.

\subsection{Generality and Platform Independence}

Although the reference implementation is built on ROS\,2 and PyBullet,
the proposed architecture is not tied to a specific middleware or simulation framework.
The separation between agent logic, capability packages, and runtime enforcement
allows the system to be mapped onto different execution backends.

In particular, the persistent agent in our architecture represents a logical
control entity rather than a specific implementation.
It can be instantiated using a wide range of agentic systems,
including LLM-based planning agents or conventional rule-based controllers,
making the framework both model-agnostic and implementation-agnostic.
Similarly, the ECM interface and runtime can be adapted to different
hardware platforms or real-time control systems.
This suggests that the architecture defines a candidate computational model
for embodied intelligent systems,
rather than a framework specific to a particular software stack.
Validating this generality across diverse platforms and task domains
remains an important direction for future work.

\subsection{Single-Agent Scope and Multi-Robot Extension}

This work focuses on the single-robot setting,
where one persistent agent governs the entire system.
We acknowledge that multi-agent decomposition within a single robot
offers genuine advantages:
concurrent reasoning across subsystems can reduce latency,
specialized agents can be independently developed and tested,
and agent boundaries provide natural fault-isolation units.
These properties are particularly valuable in large-scale systems
with heterogeneous hardware and time-critical subsystems.

However, we argue that for a broad class of manipulation and service tasks,
these benefits are outweighed by the costs of
maintaining consistency across multiple control authorities,
resolving conflicts in shared state,
and debugging emergent behavior arising from inter-agent interactions.
The single-agent model trades parallelism for coherence:
it provides a single locus of identity, memory, and decision authority
that simplifies reasoning about system behavior and safety.

Importantly, this does not preclude multi-robot scenarios.
Instead, we envision a system in which each robot corresponds to one persistent agent,
and multi-robot coordination emerges from interactions between agents at a higher level.
This distinction avoids introducing multiple agents within a single robot,
while still allowing distributed systems to be constructed
as collections of single-agent robots.

A rigorous empirical comparison between single-agent and multi-agent architectures
on the same task remains an important open question.

\paragraph{Abstraction-level objection.}
A modern AEROS agent may internally consist of multiple sub-models
(LLM planner, vision encoder, motion controller).
We adopt a principled criterion:
a component is an \emph{agent} if and only if it maintains
its own persistent identity, goal state, and decision authority.
Stateless sub-models invoked by the agent are \emph{tools}, not agents:
agents negotiate, tools are called.

\subsection{Future Directions}

Key directions include:
(1)~real-robot deployment with physical policy validation and sim-to-real transfer;
(2)~learning-based capability acquisition, where skills are learned from data
and packaged as ECMs;
(3)~large-scale capability ecosystems with ECM versioning, sharing, and trust;
and (4)~standardized architectural benchmarks enabling controlled
architecture-vs-architecture comparisons.

% ============================================================
\section{Conclusion}
\label{sec:conclusion}

This paper introduced AEROS, an architecture built on three core principles:
one robot hosts exactly one persistent agent (the Single-Agent Robot Principle),
capabilities are delivered through installable Embodied Capability Modules,
and safety is enforced by a policy-separated runtime
that is independent of capability logic.

Evaluation in PyBullet simulation across eight experiments
and three diverse tasks showed that the architecture enables
robust dynamic re\-planning (100\% success vs.\ 51\%--72\% static planning),
failure recovery under stochastic skill failures,
deterministic policy enforcement with zero false acceptances,
consistent cross-task generality without task-specific tuning,
published-baseline comparison showing statistical significance over BehaviorTree.CPP
and ProgPrompt-style execution (100\% vs.\ 92.7\%, $p<0.05$),
runtime ECM hot-swapping with negligible overhead ($<0.001$\,ms),
ablation analysis confirming dynamic re-planning and recovery as the primary
contributors to robustness,
and graceful degradation under extreme failure rates ($>$97\% success at
$p_{\text{fail}}=80\%$, 82.3\% at $p_{\text{fail}}=90\%$).
The primary limitations---simulation-only validation, reimplemented rather than
original-framework baselines, and untested memory/lifecycle features---are acknowledged
and targeted as priorities for future work, alongside
real-robot deployment and scalability analysis.

In the longer term, this architecture points toward a robotic operating model
in which capabilities can be installed, composed, and governed in a principled manner,
analogous to application ecosystems in modern computing platforms.

% ============================================================
\appendix
\section{Planning Rules}
\label{app:planning_rules}

This appendix lists the complete rule-based planning logic
used in the reference implementation evaluated in Section~7.
The planner is invoked at each re-planning cycle and returns
a task graph conditioned on the current world state.
Each task's plan skill inspects world-state predicates and
emits only the steps that remain incomplete.

\subsection*{Task Routing (Agent-Level Planner)}

The agent-level planner maps natural-language instructions to
the appropriate ECM plan skill via keyword matching:

\begin{verbatim}
if "dumpling" in instruction:
    return "dumpling.plan"
elif "clean" or "table" in instruction:
    return "clean.plan"
elif "fetch" or "bring" or "retrieve":
    return "fetch.plan"
\end{verbatim}

\subsection*{Dumpling Task Rules}

\begin{small}
\begin{verbatim}
def plan(world_state):
    steps = []
    if not world.dough_on_workspace
       or not world.filling_on_workspace:
        steps += [dumpling.prepare]
    if not world.wrapper_aligned
       and not world.dumpling_wrapped:
        steps += [dumpling.recover]
    if not world.dumpling_wrapped:
        steps += [dumpling.wrap,
                  retry=2,
                  on_failure=dumpling.recover]
    if not world.dumpling_cooked:
        steps += [dumpling.boil]
    return steps
\end{verbatim}
\end{small}

Four world-state predicates (\texttt{dough\_on\_workspace},
\texttt{wrapper\_aligned}, \texttt{dumpling\_\allowbreak wrapped},
\texttt{dumpling\_\allowbreak cooked}) determine which steps are emitted.
The \texttt{wrap} step includes a retry budget of~2 and a
recovery fallback (\texttt{dumpling.recover}) that realigns the wrapper.

\subsection*{Clean Table Task Rules}

\begin{small}
\begin{verbatim}
def plan(world_state):
    steps = []
    if not world.table_wiped:
        steps += [clean.wipe]
    if not world.table_organized:
        steps += [clean.organize]
    return steps
\end{verbatim}
\end{small}

Two predicates (\texttt{table\_wiped}, \texttt{table\_organized}).
Retry and recovery for \texttt{clean.wipe} are injected at the
agent level (retry=1, on\_failure=clean.recover).

\subsection*{Fetch Object Task Rules}

\begin{small}
\begin{verbatim}
def plan(world_state):
    steps = []
    if not world.robot_at_target:
        steps += [fetch.navigate]
    if not world.object_detected:
        steps += [fetch.detect]
    if not world.object_grasped:
        steps += [fetch.grasp,
                  retry=1,
                  on_failure=fetch.recover]
    if not world.object_delivered:
        steps += [fetch.deliver]
    return steps
\end{verbatim}
\end{small}

Four predicates spanning navigation and manipulation.
The \texttt{grasp} step includes retry and recovery to handle
stochastic perception failures (35\% failure rate in evaluation).

\subsection*{Generality Assessment}

The planning rules are strictly world-state-conditioned:
each rule checks a single Boolean predicate and emits
the corresponding skill if the predicate is unsatisfied.
This pattern generalizes to any task expressible as a
sequence of state transitions, provided the task's
completion conditions can be represented as
world-state predicates.
The rules are intentionally simple to isolate
architectural contributions from planner sophistication,
as discussed in Section~6.4.

\bibliographystyle{elsarticle-harv}
\bibliography{references}

\end{document}